\documentclass[10pt,twocolumn,letterpaper]{article}

\usepackage[pagenumbers]{cvpr} 

\usepackage{amsmath,amsfonts,bm}

\def\eqref#1{equation~\ref{#1}}

\def\1{\bm{1}}

\DeclareMathAlphabet{\mathsfit}{\encodingdefault}{\sfdefault}{m}{sl}
\SetMathAlphabet{\mathsfit}{bold}{\encodingdefault}{\sfdefault}{bx}{n}

\usepackage{graphicx}
\usepackage{amsmath}
\usepackage{amssymb}
\usepackage{booktabs}
\usepackage{algorithm}
\usepackage{dsfont}
\usepackage{algpseudocode}
\usepackage{algorithmicx}
\usepackage{mathtools}
\usepackage{multirow}
\usepackage{graphics}
\usepackage{wrapfig}
\usepackage{bbm}
\usepackage{xspace}
\usepackage{xcolor}
\usepackage{tikz}
\usepackage{adjustbox}
\usepackage{pifont}
\usepackage[accsupp]{axessibility}  

\newcommand{\methodName}{RIST\xspace} 

\newcommand{\cmark}{\ding{51}}%
\usepackage{xr}
\makeatletter
\newcommand*{\addFileDependency}[1]{
  \typeout{(#1)}
  \@addtofilelist{#1}
  \IfFileExists{#1}{}{\typeout{No file #1.}}
}
\makeatother

\definecolor{cvprblue}{rgb}{0.21,0.49,0.74}
\usepackage[pagebackref,breaklinks,colorlinks,citecolor=cvprblue]{hyperref}

\def\paperID{8074} 
\def\confName{CVPR}
\def\confYear{2024}

\title{Learning SO(3)-Invariant Semantic Correspondence via Local Shape Transform}

\author{
Chunghyun Park$^{1}$
\hspace{-1.4mm}\thanks{Equal contribution}
\hspace{8mm}
Seungwook Kim$^{1}$\hspace{-0.35mm}\footnotemark[1]
\hspace{8mm}
Jaesik Park$^2$
\hspace{8mm}
Minsu Cho$^{1}$\vspace{2mm}\\
$^1$POSTECH
\hspace{10mm}
$^2$Seoul National University\vspace{2mm}\\
}

\begin{document}
\maketitle

\begin{abstract}

Establishing accurate 3D correspondences between shapes stands as a pivotal challenge with profound implications for computer vision and robotics.
However, existing self-supervised methods for this problem assume perfect input shape alignment, restricting their real-world applicability.
In this work, we introduce a novel self-supervised \textbf{R}otation-\textbf{I}nvariant 3D correspondence learner with local \textbf{S}hape \textbf{T}ransform, dubbed \methodName, that learns to establish dense correspondences between shapes even under challenging intra-class variations and arbitrary orientations.
Specifically, \methodName learns to dynamically formulate an SO(3)-invariant local shape transform for each point, which maps the SO(3)-equivariant global shape descriptor of the input shape to a local shape descriptor.
These local shape descriptors are provided as inputs to our decoder to facilitate point cloud self- and cross-reconstruction.
Our proposed self-supervised training pipeline encourages semantically corresponding points from different shapes to be mapped to similar local shape descriptors, enabling \methodName to establish dense point-wise correspondences.
\methodName demonstrates state-of-the-art performances on 3D part label transfer and semantic keypoint transfer given arbitrarily rotated point cloud pairs of the same category, outperforming existing methods by significant margins.

\end{abstract}

\section{Introduction}
\label{sec:intro}

Establishing dense 3D correspondences between different shapes is foundational to numerous applications across computer vision, graphics, and robotics~\cite{saxena2006robotic, miller2003automatic, hao2013efficient, zeng20203d}.
One of the primary challenges hindering advancements in this domain is the difficulty of annotating dense inter-shape correspondences, which limits the leverage of strongly-supervised learning paradigms.

\begin{figure}
    \centering
    \includegraphics[width=\linewidth]{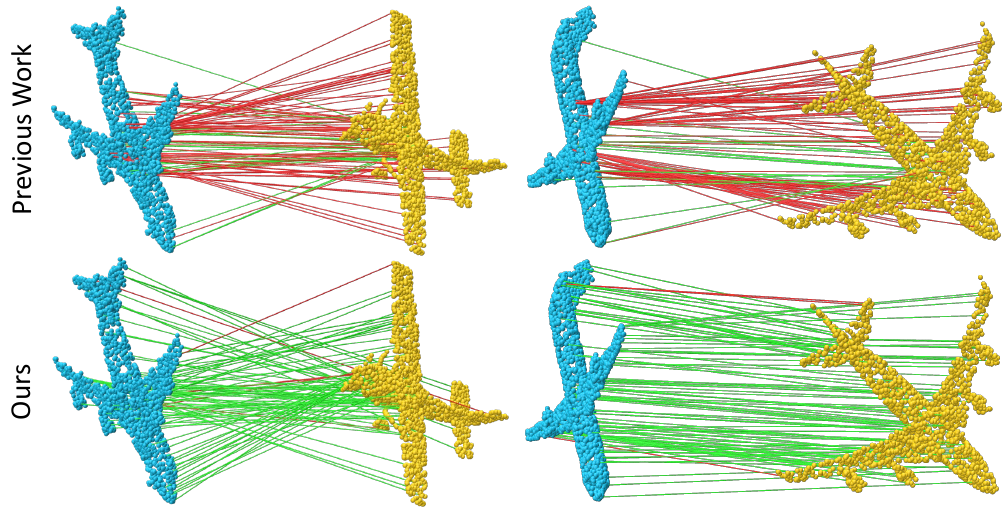}
    \caption{\textbf{Semantic correspondence between rotated shapes.} 
    We visualize the semantic correspondence results of the previous SOTA~\cite{cheng2021learning} and ours, given two randomly rotated airplanes from the ShapeNetPart dataset~\cite{shapenetpart}.
    Green and red lines indicate the correct and incorrect matches, respectively.
    For each method, 100 source points are randomly selected from the source (yellow) for correspondence visualization\protect\footnotemark.
    Ours predicts SO(3)-invariant correspondences, showing superior accuracy and robustness in comparison to the previous SOTA~\cite{cheng2021learning}.}
    \vspace{-4mm}
    \label{fig:teaser}
\end{figure}

Recently, self-supervised learning methods have been proposed to address this issue~\cite{liu2020learning,cheng2021learning}, showing promising directions for 3D correspondence estimation.
Nonetheless, a significant limitation in existing approaches is their stringent assumption about the alignment of input shape pairs; these methods strongly assume that the input point cloud pair to establish correspondences between is precisely aligned.
This assumption is rarely met in practice, where object scans and shape instances can be arbitrarily oriented.
We find that the performance of existing methods degrades significantly when confronted with rotated input shapes, restricting their real-world applicability (Figure~\ref{fig:teaser}).

\footnotetext{For the details of the inference algorithm, please refer to Appendix~\ref{sec:supp_algo}.}

To address this challenge, we introduce a novel self-supervised learning approach, dubbed \methodName, designed to reliably determine dense \textit{SO(3)-invariant} correspondences between shapes via local shape transform.
In essence, \methodName learns to formulate SO(3)-invariant local shape transform for each point in a dynamic and input-dependent manner.
Each point-wise local shape transform maps the SO(3)-equivariant \textit{global} shape descriptor of the input shape to a \textit{local} shape descriptor, which is passed to the decoder to reconstruct the shape and pose of the input shape.
By training \methodName via self- and cross-reconstruction of input shapes, true semantically corresponding points are trained to yield similar local shape descriptors, enabling us to determine dense shape correspondences.

\methodName demonstrates state-of-the-art performance on part segmentation label transfer when evaluated on the ShapeNetPart~\cite{shapenetpart} and ScanObjectNN~\cite{scanobjectnn} datasets.
In particular, significant improvements over existing baselines are observed when our method is applied to randomly oriented shape pair inputs.
Furthermore, our approach also proves to be more effective compared to existing methods at semantic keypoint transfer when evaluated on the KeypointNet dataset~\cite{you2020keypointnet}.
This showcases not only the applicability of \methodName across a diverse range of tasks, but also its potential to be utilized for efficient dense annotation of 3D shapes.
These results highlight the efficacy of \methodName in addressing the challenges posed by real-world scenarios where existing methods fail to perform effectively.

The main contributions of our work are as follows:
\begin{itemize}
\item We introduce \methodName, a novel self-supervised approach for determining dense SO(3)-invariant correspondences between arbitrarily aligned 3D objects. 
\item We propose to formulate the local shape information of each point as a novel function called \textit{local shape transform} with dynamic input-dependent parameters, which effectively maps the global shape descriptor of input shapes to local shape descriptors.
\item \methodName achieves state-of-the-art performance on 3D part segmentation label transfer and 3D keypoint transfer under arbitrary rotations, indicating its potential for application in a wide range of practical tasks in computer vision.
\end{itemize}
\section{Related Work}
\label{sec:related_work}

\noindent
\textbf{Point cloud understanding via self-supervised learning.}
While traditional methods for point cloud processing involving hand-crafted features~\cite{tombari2010unique, salti2014shot} have shown impressive performance, with the advent of deep learning, substantial research efforts have been directed towards developing learning-based algorithms capable of effectively processing and understanding point clouds~\cite{qi2017pointnet, qi2017pointnetplusplus, zhao2021point, park2022fast, choe2022pointmixer}. 
Due to limited large-scale datasets with rich annotations, self-supervised learning approaches emerged as a viable alternative.
One of the most prominent directions to learn point cloud representations in a self-supervised manner is learning through self-reconstruction~\cite{yang2018foldingnet, zhao20193d, pang2022masked} of the point cloud.
Primarily inspired by the efficacy of point cloud reconstruction as a self-supervised representation learning scheme, we train \methodName to establish 3D correspondences in a self-supervised manner via self- and cross-reconstruction of point clouds by leveraging SO(3)-invariant dynamic local shape transform.


\vspace{2mm}
\noindent
\textbf{Equivariance and invariance to rotation.}
The conventional method to improve a neural network's robustness to rotation is by employing rotation augmentations during training or inference.
However, this tends to increase the resources required for training and still shows unsatisfactory results when confronted with an unseen rotation~\cite{Li_2021_ICCV, kim2023stable}.
In recent years, various methods have been proposed to yield point cloud representations, which are equivariant~\cite{cohen2018spherical, thomas2018tensor, shen20203d, chen2021equivariant} or invariant~\cite{sun2019srinet, li2021rotation, xiao2021triangle, Li_2021_ICCV, kim2023stable} to the rotation of the input, demonstrating enhanced performances under arbitrary input rotations.
To facilitate the rotation-robust establishment of 3D dense correspondences, we utilize SO(3)-equivariant networks
in building \methodName, leveraging SO(3)-equivariant and -invariant representations to guarantee robustness to rotation by design.

\vspace{2mm}
\noindent
\textbf{Semantic correspondences under intra-class variations.}
Finding correspondences between images or shapes under intra-class variations - manifesting as differences in shape, size, and orientation within the same category of objects - poses significant challenges over photometric or viewpoint variations.
This task has been widely studied in the domain of images, where existing methods make use of sparsely annotated image pair datasets to train their method in a strongly- or a weakly-supervised manner~\cite{cho2021cats,kim2022transformatcher, truong2022probabilistic, huang2022learning, kim2024efficient}.
However, learning to establish dense yet reliable 3D correspondences between 3D shapes remains challenging, as it is infeasible to label dense correspondence annotations across point cloud pairs with intra-class variations.
Self-supervised methods have been proposed to address this issue~\cite{liu2020learning, cheng2021learning}, but they strongly assume that the input point clouds are aligned, leading to considerable significant degradation when confronted with arbitrarily rotated point clouds.
Additionally, the functional map-based approach introduced by Huang~\etal~\cite{huang2022multiway} for non-rigid registration struggles with topological changes and efficiency.
To this end, we propose \methodName to establish reliable 3D dense correspondences irrespective of the input point clouds' poses. 

\begin{figure*}[ht]
    \begin{center}
        \includegraphics[width=\linewidth]{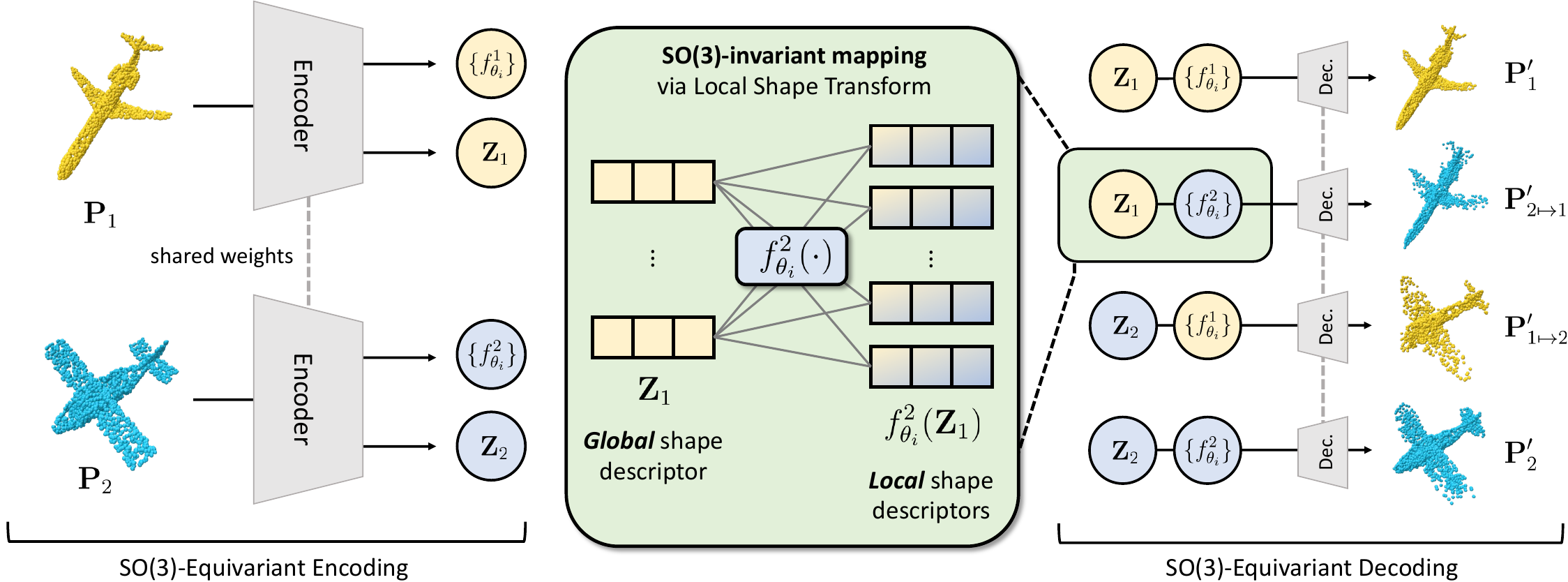}
    \end{center}
      \caption{\textbf{Overview: Self-supervised training of \methodName}. 
      The input point clouds are independently encoded to SO(3)-equivariant global shape descriptor $\mathbf{Z}$ and dynamic SO(3)-invariant point-wise local shape transforms $\{f_{\theta_i}\}$.
      The local shape transforms map the global shape descriptor to local shape descriptors by infusing local semantics and geometry, which are used as inputs to the decoder for self-reconstruction.
      For cross-reconstruction, we apply the local shape transforms formulated from \textit{another} point cloud to reconstruct the point cloud, ensuring that the local shape descriptors successfully capture generalizable local semantics and geometries.
      We supervise \methodName via penalizing errors in self- and  cross-reconstructions.
      At inference, we can leverage the local shape transforms for obtaining local shape descriptors, to identify the dense correspondences.
}
\vspace{-4mm}
\label{fig:main_overview}
\end{figure*}

\section{\methodName for 3D Semantic Correspondence}
\label{sec:method}

In this section, we detail the components of \methodName, which come together to facilitate the end-to-end self-supervised training for 3D semantic correspondence establishment.
The objective of 3D semantic correspondence is as follows; given two different point clouds instances $\mathbf{P}_1 \in \mathbb{R}^{N \times 3}$ and $\mathbf{P}_2 \in \mathbb{R}^{N \times 3}$ belonging to the same semantic category, we aim to find all semantically corresponding point pairs $\{\mathbf{p}_i, \mathbf{q}_i\}_{i=1}^{N'}$\footnote{$N'\leq N$; 
there could be points with no matches.} such that $\mathbf{p}_i \in \mathbf{P}_1$ and $\mathbf{q}_i \in \mathbf{P}_2$.
To achieve this, we claim it is crucial to identify the local shape information \ie local semantics and geometry, which is generalizable across different instances within the same category.

Therefore, the main idea of \methodName is to dynamically generate a SO(3)-invariant local shape transform as a function for each point, such that each local shape transform can map the SO(3)-equivariant global shape descriptor of the input point cloud to its respective local shape descriptor.
In the following, we elaborate on the network architecture of \methodName, in particular how we leverage SO(3)-equivariant and invariant representations to facilitate the dynamic formulation of pointwise SO(3)-invariant local shape transforms and the reconstruction of pose-preserved point clouds (\cref{subsec:architecture}).
Subsequently, we introduce our self-supervisory objective function, which trains \methodName to self- and cross-reconstruct the input point clouds in a rotation-equivariant manner (\cref{subsec:loss}), finally enabling the establishment of 3D dense correspondences (\cref{subsec:matching}) via corresponding local shape descriptors.
Figure~\ref{fig:main_overview} illustrates the outline of the training scheme of \methodName.

\subsection{Network Design of \methodName}
\label{subsec:architecture}
 
\subsubsection{Preliminary: SO(3)-Equivariant Representation}
\label{subsubsec:preliminary}
One of the main motivations of \methodName is to establish reliable and accurate 3D dense correspondences given \textit{arbitrarily rotated} shapes, a setting where existing work shows to be brittle.
This requires our encoder to formulate the point-wise local shape transforms not only effectively to capture the local shape semantics and geometry, but also robustly against transformations in the SO(3) space.
To this end, we integrate SO(3)-equivariant networks into \methodName to facilitate robustness to SO(3) transformations of the input.
In this work, we choose VNNs~\cite{deng2021vector} to build our SO(3)-equivariant layers for their simplicity and efficacy.
In VNNs, a single neuron, which is represented by a scalar-list of values, is lifted to a \textit{vector-list} feature $\mathbf{V} \in \mathbb{R}^{C \times 3}$, which is essentially a sequence of 3D vectors.
The layers of VNNs handle batches of such vector-list features such that equivariance with respect to rotation $R \in$ SO(3) is satisfied \ie $f(\mathbf{V}R) = f(\mathbf{V})R$\footnote{We refer the readers to the original paper~\cite{deng2021vector} for further information and detailed formulations of VNNs.}.
Notably, we can yield an SO(3)-\textit{invariant} output by performing a product of an equivariant vector-list feature $\mathbf{V}R \in \mathbb{R}^{C \times 3}$ with the transpose of another consistently equivariant vector-list feature $\mathbf{U}R \in \mathbb{R}^{C' \times 3}$ as follows: $(\mathbf{V}R)(\mathbf{U}R)^\top = \mathbf{V}RR^\top\mathbf{U}^\top=\mathbf{V}\mathbf{U}^\top$.
This serves as a critical functionality when constructing our SO(3)-invariant local shape transform of our encoder (\cref{subsubsec:encoder}).

\subsubsection{SO(3)-Equivariant Encoder}
\label{subsubsec:encoder}
We design our encoder architecture to take as input a point cloud $\mathbf{P} \in \mathbb{R}^{N \times 3}$, and simultaneously output an SO(3)-\textit{equivariant} global shape descriptor and formulate point-wise SO(3)-\textit{invariant} local shape transforms.

\smallbreak
\noindent
\textbf{SO(3)-equivariant global shape descriptor.}
Given a point cloud, we first aim to obtain the SO(3)-equivariant \textit{global} shape descriptor $\mathbf{Z} \in \mathbb{R}^{C \times 3}$, which captures the pose and the global shape characteristics of the input point cloud.
We leverage VN-DGCNN~\cite{deng2021vector} as our encoder architecture, which consists of 4 edge convolutional VN-layers to capture local semantics at a progressively larger receptive field, and a FPN~\cite{lin2017feature} to aggregate the multi-level features.
Then, we apply the global average pooling to the aggregated SO(3)-equivariant point-wise features $\mathbf{V}^{\mathrm{equi}} \in \mathbb{R}^{C\times 3 \times N}$ to encode SO(3)-equivariant global shape descriptor $\mathbf{Z}$ of the input point cloud.
The global shape descriptor can be used subsequently as the input for our SO(3)-invariant point-wise local shape transform, to be mapped to their respective local shape descriptors, as shown in Figure~\ref{fig:main_overview}. 

\smallbreak
\noindent
\textbf{SO(3)-invariant local shape transform.}
Alongside the extraction of SO(3)-equivariant global shape descriptors, we also formulate the SO(3)-\textit{invariant} local shape transform $f_{\theta_i}: \mathbb{R}^{C \times 3} \mapsto \mathbb{R}^{C' \times 3}$ for each point $\mathbf{p}_i\in\mathbb{R}^3$ of the input point cloud $\mathbf{P} \in \mathbb{R}^{N\times 3}$.
The parameters of each local shape transform $\theta_i \in \mathbb{R}^{C' \times C}$ are input-dependent - thus, dynamic since they are predicted by our encoder for the $i$-th point of the point cloud. 
To predict $\theta_i$, we first obtain SO(3)-\textit{invariant} point-wise features $\mathbf{V}^{\mathrm{in}} \in \mathbb{R}^{C' \times 3 \times N}$ as described in Sec.~\ref{subsubsec:preliminary}.
Then, we transform each vectorized SO(3)-invariant point-wise feature $\operatorname{vec}(\mathbf{v}^{\mathrm{in}}_i) \in \mathbb{R}^{3C'}$ to the vectorized parameter of the local shape transform $\operatorname{vec}(\theta_i) \in \mathbb{R}^{C'C}$ by using a multi-layer perceptron.
By reshaping $\operatorname{vec}(\theta_i)$ to $\theta_i \in \mathbb{R}^{C'\times C}$, we finally obtain the dynamic and SO(3)-invariant local shape transform $f_{\theta_i}$ for the point $\mathbf{p}_i$.

The role of these local shape transforms is to map the SO(3)-equivariant global shape descriptor $\mathbf{Z} \in \mathbb{R}^{C \times 3}$ to their respective local shape descriptors $\mathbf{v}'_i:=f_{\theta_i}(\mathbf{Z}) \in \mathbb{R}^{C' \times 3}$, which is provided as the input to our decoder for reconstruction.
Our self-supervised training scheme encourages the point-wise dynamic local shape transform to encapsulate the local shape information \eg semantics and geometry, to enhance the reconstruction performance.

\subsubsection{SO(3)-Equivariant Decoder} 
Our decoder aims to reconstruct the initial input shapes using the obtained SO(3)-equivariant global shape descriptors $\mathbf{Z}$ and the SO(3)-invariant local shape transforms $\{f_{\theta_i}\}_{i=1}^N$.
To reconstruct the point clouds aligned to their initial poses, we leverage SO(3)-equivariant layers as the building blocks of our decoder architecture.
We first train our decoder to perform self-reconstruction, using the local shape descriptors $\mathbf{V}'$, \ie $\mathbf{P} \leftrightarrow \mathbf{P}':= \operatorname{Decoder}(\mathbf{V}') = \operatorname{Decoder}(\{f_{\theta_i}(\mathbf{Z})\}_{i=1}^{N})$.
We also train our decoder to perform cross-reconstruction, where we use the local shape descriptors obtained using global shape descriptors and local shape transforms from \textit{different} point clouds.
Specifically, assume we are given two point clouds $\mathbf{P}_1,\mathbf{P}_2 \in \mathbb{R}^{N \times 3}$, with SO(3)-equivariant global shape descriptors $\mathbf{Z}_1, \mathbf{Z}_2 \in \mathbb{R}^{C \times 3}$ and SO(3)-invariant local shape transforms $\{f_{\theta_i}^1\}_{i=1}^N, \{f_{\theta_i}^2\}_{i=1}^N$.
As shown in Figure~\ref{fig:main_overview}, we then can perform cross-reconstruction from $\mathbf{P}_2$ to $\mathbf{P}_1$ as follows: $\mathbf{P}_1 \leftrightarrow \mathbf{P}'_{2\mapsto 1}:=\operatorname{Decoder}(\{f_{\theta_i}^2(\mathbf{Z_1})\}_{i=1}^N)$.
Intuitively, for the above cross-reconstruction to be carried out successfully, the local shape transforms for points of a \textit{true} correspondence should hold similar dynamic parameters, mapping global shape descriptors to similar local shape descriptors.
By training \methodName to cross-reconstruct point clouds, we are supervising local shape transforms to map corresponding points between shapes to similar local shape descriptors, which encode local semantics and geometry that are generalizable across different instances within a category.

\subsection{Self-Supervised Objective}
\label{subsec:loss}

Due to the lack of annotated datasets for dense 3D inter-shape correspondences, we train \methodName in a self-supervised manner by penalizing inaccurate shape reconstructions.
First, we supervise \methodName for self-reconstruction using the following loss:
\begin{equation}
    \label{eq:sr}
    \mathcal{L}_\mathrm{SR} = \lambda_{\mathrm{MSE}}\operatorname{MSE}(\mathbf{P}, \mathbf{P}') + \lambda_{\mathrm{EMD}}\operatorname{EMD}(\mathbf{P}, \mathbf{P}'),
\end{equation}
where $\operatorname{MSE}$ is the Mean Squared Error, $\operatorname{EMD}$ stands for the Earth Mover's Distance, and both $\lambda_{\mathrm{MSE}}$ and $\lambda_{\mathrm{EMD}}$ are weight coefficients. 
In essence, we are trying to minimize the difference between the input and reconstructed point cloud.
We also supervise \methodName for cross-reconstruction as follows:
$\mathcal{L}_\mathrm{CR} = \lambda_{\mathrm{CD}}\operatorname{CD}(\mathbf{P}_1, \mathbf{P}'_{2 \mapsto 1})$, where $\operatorname{CD}$ stands for the Chamfer distance, and $\lambda_{\mathrm{CD}}$ is a weight coefficient.
Finally, our total loss $\mathcal{L}_{\mathrm{total}}$ is defined as: $\mathcal{L}_{\mathrm{total}} = \mathcal{L}_{\mathrm{SR}} + \mathcal{L}_{\mathrm{CR}}$.
We omit the $\operatorname{CD}$ loss from self-reconstruction, as we can directly use the input point cloud to provide supervision using the $\operatorname{MSE}$ loss.
We also omit the $\operatorname{EMD}$ loss from cross-reconstruction, as $\operatorname{EMD}$ tends to overlook the fidelity of detailed structures~\cite{wu2021density}, which is crucial in cross-reconstruction of shapes under intra-class variations.

\subsection{SO(3)-Invariant Correspondence}
\label{subsec:matching}
In this section, given two randomly rotated point clouds $\mathbf{P}_1$ and $\mathbf{P}_2$, we elaborate on how our \methodName establishes the 3D dense correspondence from $\mathbf{P}_1$ to $\mathbf{P}_2$. As shown in Figure~\ref{fig:main_overview}, we first encode the SO(3)-equivariant global shape descriptor of $\mathbf{P}_1$, $\mathbf{Z}_1 \in \mathbb{R}^{C\times3}$, and the SO(3)-invariant local shape transform functions of $\mathbf{P}_2$, $\{f_{\theta_i}^2\}_{i=1}^N$. Then, we cross-reconstruct $\mathbf{P}_1$ as follows: $\mathbf{P}'_{2\mapsto 1}:=\operatorname{Decoder}(\{f_{\theta_i}^2(\mathbf{Z_1})\}_{i=1}^N)$. Finally, we define the 3D dense correspondence from $\mathbf{P}_2$ to $\mathbf{P}_1$ as the nearest point pairs among all possible pairs between $\mathbf{P}_1$ and $\mathbf{P}_{2\mapsto1}'$. Since both encoder and decoder are SO(3)-equivariant, the cross-reconstructed point cloud $\mathbf{P}_{2\mapsto1}'$ is aligned to $\mathbf{P}_1$. As a result, our \methodName can predict 3D dense correspondences between randomly rotated point clouds, while previous approaches~\cite{cheng2021learning, liu2020learning} experience a high rate of failure.


\section{Experiments}
\label{sec:experiments}

\begin{table*}[t]
    \centering
    \begin{adjustbox}{width=0.9\textwidth,center}
       \begin{tabular}{clccccccccc}
                \toprule
                Training & Method & Airplane & Cap & Chair & Guitar & Laptop & Motorcycle & Mug & Table & Average \\                
                \midrule
                \multirow{5}{*}[-0.5ex]{\shortstack{w/o \\ Rotations}} & FoldingNet~\cite{yang2018foldingnet}
                                   & 17.8 & 34.7 & 22.5 & 22.1 & \underline{36.2} & 12.6 & 50.0 & 34.6 & 28.8 \\
                & AtlasNetV2~\cite{deprelle2019learning} 
                                   & 19.7 & 31.4 & 23.6 & 22.7 & 36.0 & 13.1 & 49.7 & 35.2 & 28.9 \\
                & DPC~\cite{lang2021dpc} 
                                   & \underline{22.7} & 37.1 & 25.6 & \underline{31.9} & 35.0 & \underline{17.5} & 51.3 & \underline{36.8} & \underline{32.2} \\
                & CPAE~\cite{cheng2021learning} 
                                   & 21.0 & \underline{38.0} & \underline{26.0} & 22.7 & 34.9 & 14.7 & \underline{51.4} & 35.5 & 30.5 \\
                & \methodName (ours) & \textbf{52.1} & \textbf{54.5} & \textbf{58.3} & \textbf{74.1} & \textbf{56.5} & \textbf{48.6} & \textbf{75.0} & \textbf{41.3} & \textbf{57.6} \\
                \midrule
                \multirow{5}{*}[-0.5ex]{\shortstack{w/ \\ Rotations}} & FoldingNet~\cite{yang2018foldingnet}
                                   & 22.5 & 33.2 & 24.0 & 31.0 & 35.9 & 13.5 & 49.9 & 37.0 & 30.9 \\
                & AtlasNetV2~\cite{deprelle2019learning} 
                                   & 21.1 & 32.7 & 25.2 & 28.8 & 35.5 & 14.5 & 49.9 & \underline{41.0} & 31.1 \\
                & DPC~\cite{lang2021dpc} 
                                   & \underline{24.6} & \underline{38.5} & \underline{25.6} & \underline{40.2} & 34.9 & \underline{19.3} & 51.8 & 37.3 & \underline{34.0} \\
                & CPAE~\cite{cheng2021learning} 
                                   & 17.0 & 36.6 & 24.5 & 39.4 & \underline{37.4} & 15.8 & \underline{51.9} & 36.7 & 32.4 \\
                & \methodName (ours) & \textbf{51.2} & \textbf{57.0} & \textbf{55.0} & \textbf{73.5} & \textbf{60.6} & \textbf{48.5} & \textbf{72.2} & \textbf{44.4} & \textbf{57.8} \\
               
                \bottomrule
        \end{tabular}
        \end{adjustbox}
        \caption{\textbf{Average IoU (\%) of part label transfer for eight categories in the ShapeNetPart dataset~\cite{shapenetpart}.} 
        Ours consistently outperforms previous approaches~\cite{yang2018foldingnet, deprelle2019learning, lang2021dpc, cheng2021learning} both with and without rotation augmentations during training, achieving the state-of-the-art IoU. We also provide results of the other classes in Appendix~\ref{sec:supp_partlabeltransfer}.
        }
        \vspace{-4mm}
        \label{tbl:part_label_transfer}
\end{table*}

                


        

We present evaluations of \methodName on the tasks of 3D part segmentation label transfer and 3D semantic keypoint transfer, following prior work~\citep{liu2020learning, cheng2021learning}.
For both tasks, each method is trained with and without rotation augmentations, and tested on arbitrarily rotated inputs to validate its rotation robustness in predicting semantic correspondences between the rotated inputs.
Note that previous approaches~\cite{yang2018foldingnet, deprelle2019learning, liu2020learning, cheng2021learning} used aligned point clouds or slightly rotated point clouds with a subset of SO(3), not the full SO(3), for testing - which is unrealistic in practice, as described in Appendix~\ref{sec:supp_so3}.


\noindent
\textbf{Datasets.}
We use the ShapeNetPart~\cite{shapenetpart} and ScanObjectNN~\cite{scanobjectnn} datasets to evaluate \methodName on the task of 3D part segmentation label transfer, which requires 3D dense semantic correspondence. 
The ShapeNetPart dataset~\cite{shapenetpart} consists of 16,880 synthetic 3D data from 16 categories.
The ScanObjectNN dataset~\cite{scanobjectnn} contains \textit{real-world} scanned data, and provides part label annotations for the chair category.
Following the previous work~\cite{cheng2021learning}, we use the same pre-processed KeypointNet~\cite{you2020keypointnet} dataset for the 3D semantic keypoint transfer task.
Since both KeypointNet and ShapeNetPart are based on the ShapeNet dataset~\cite{chang2015shapenet}, we use the eight overlapping categories between the ShapeNetPart and KeypointNet~\cite{you2020keypointnet} datasets to evaluate each method on both tasks without fine-tuning the method on each dataset.
For all tasks, we follow the experiment setting of the previous work~\cite{cheng2021learning}.


\noindent
\textbf{Baseline methods.}
Throughout the evaluation section, we mainly compare \methodName against CPAE~\citep{cheng2021learning}, the state-of-the-art self-supervised method to establish 3D dense correspondence by exploiting an intermediate UV canonical space.
When open-sourced pre-trained models or codes are applicable, we also compare \methodName with AtlasNetV2~\citep{deprelle2019learning}, FoldingNet~\citep{yang2018foldingnet} and DPC~\cite{lang2021dpc}.
AtlasNetV2 proposes to represent shapes as the deformation and combination of learnable elementary 3D structures, which can be extended to 3D correspondence establishment.
FoldingNet introduces a folding-based decoder to `fold' a canonical 2D grid into the 3D object surface, where the canonical 2D grid can be applied to identify cross-shape correspondences.
DPC predicts 3D dense correspondence between non-rigidly deformed 3D humans, meaning that it can be a powerful baseline for both 3D part segmentation label transfer and 3D keypoint transfer tasks as well.


\noindent
\textbf{Implementation details.}
We use VN-DGCNN~\citep{deng2021vector} as our SO(3)-equivariant encoder, and VN-based multi-layer perception as our SO(3)-equivariant decoder. For a fair comparison, we set the dimension, $C$, of SO(3)-equivariant global shape descriptor $\mathbf{Z}\in\mathbb{R}^{C\times3}$ as 170 ($\approx 512 / 3$) since CPAE~\citep{cheng2021learning} uses 512-dimensional global shape descriptors.
Following the training setup of CPAE~\citep{cheng2021learning}, we use $\lambda_{\mathrm{MSE}}$, $\lambda_{\mathrm{EMD}}$, and $\lambda_{\mathrm{CD}}$ as 1000, 1, and 10, respectively.
\methodName is implemented in PyTorch, and is optimized with the Adam~\citep{kingma2014adam} optimizer at a constant learning rate of 1$e^{-3}$. 


\begin{figure*}[t]
    \begin{center}
        \includegraphics[width=0.9\linewidth]{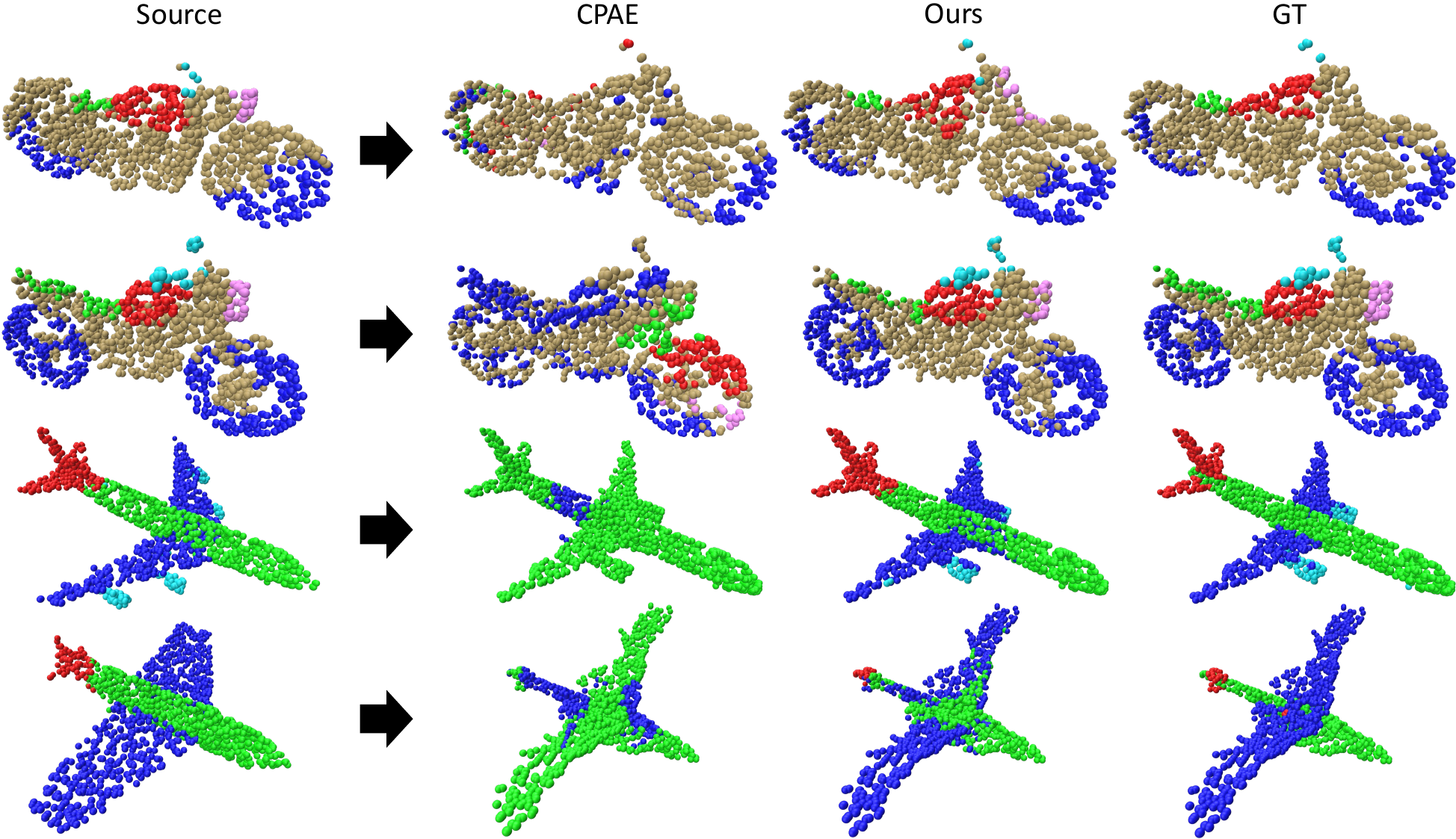}
    \end{center}
      \caption{\textbf{Qualitative results of part label transfer on the ShapeNetPart dataset~\cite{shapenetpart}.} We visualize the label transfer results via learned correspondences of each method with the ground truth labels of targets.
      Note that the input shapes were arbitrarily rotated at evaluation, differently for both the source and targets of each row, but have been aligned in the above figure for better visibility of part label transfer results.
      \methodName shows to outperform CPAE~\citep{cheng2021learning} consistently, showing a high resemblance to ground truth results.
}
\vspace{-2mm}
\label{fig:qual_part_label_transfer}
\end{figure*}
\begin{table}[t]
    \centering
    \begin{adjustbox}{width=0.8\linewidth,center}
       \begin{tabular}{lcc}
                \toprule
                Method & w/o Rotations & w/ Rotations \\               
                \midrule
                FoldingNet~\cite{yang2018foldingnet} & 23.2 & 23.3 \\
                AtlasNetV2~\cite{deprelle2019learning} & 23.6 & \underline{24.1} \\
                DPC~\cite{lang2021dpc} & 23.9 & 23.9 \\
                CPAE~\cite{cheng2021learning} & \underline{24.4} & 23.9 \\
                \methodName (ours) & \textbf{39.6} & \textbf{37.9} \\
                \bottomrule
        \end{tabular}
        \end{adjustbox}
        \caption{\textbf{Average IoU (\%) of part label transfer for the chair category in the ScanObjectNN dataset~\cite{scanobjectnn}.} Ours shows the best IoU both with and without rotation augmentations during training.
        }
        \vspace{-4mm}
        \label{tbl:part_label_transfer_scanobjectnn}
\end{table}

\subsection{Part Segmentation Label Transfer}
We compare \methodName with the state of the art in 3D part segmentation label transfer on ShapeNetPart~\cite{shapenetpart} and ScanObjectNN~\cite{scanobjectnn}. For both datasets, we use the average of instance-wise IoU scores as the evaluation metric.

\noindent \textbf{ShapeNetPart~\cite{shapenetpart}.}
The quantitative results are presented in Table~\ref{tbl:part_label_transfer}, where \methodName outperforms CPAE~\cite{cheng2021learning}, DPC~\cite{lang2021dpc}, AtlasNetV2~\cite{deprelle2019learning}, and FoldingNet~\cite{yang2018foldingnet} on all classes by a large margin with and without rotation augmentations during training.
We also provide the qualitative results of the part segmentation label transfer experiments on ShapeNetPart in Figure~\ref{fig:qual_part_label_transfer}. 
Attributing to the SO(3)-invariant nature of correspondences established by \methodName, we are able to transfer part labels significantly more accurately given randomly rotated shape pairs.

\begin{figure}[t]
    \centering
    \includegraphics[width=\linewidth]{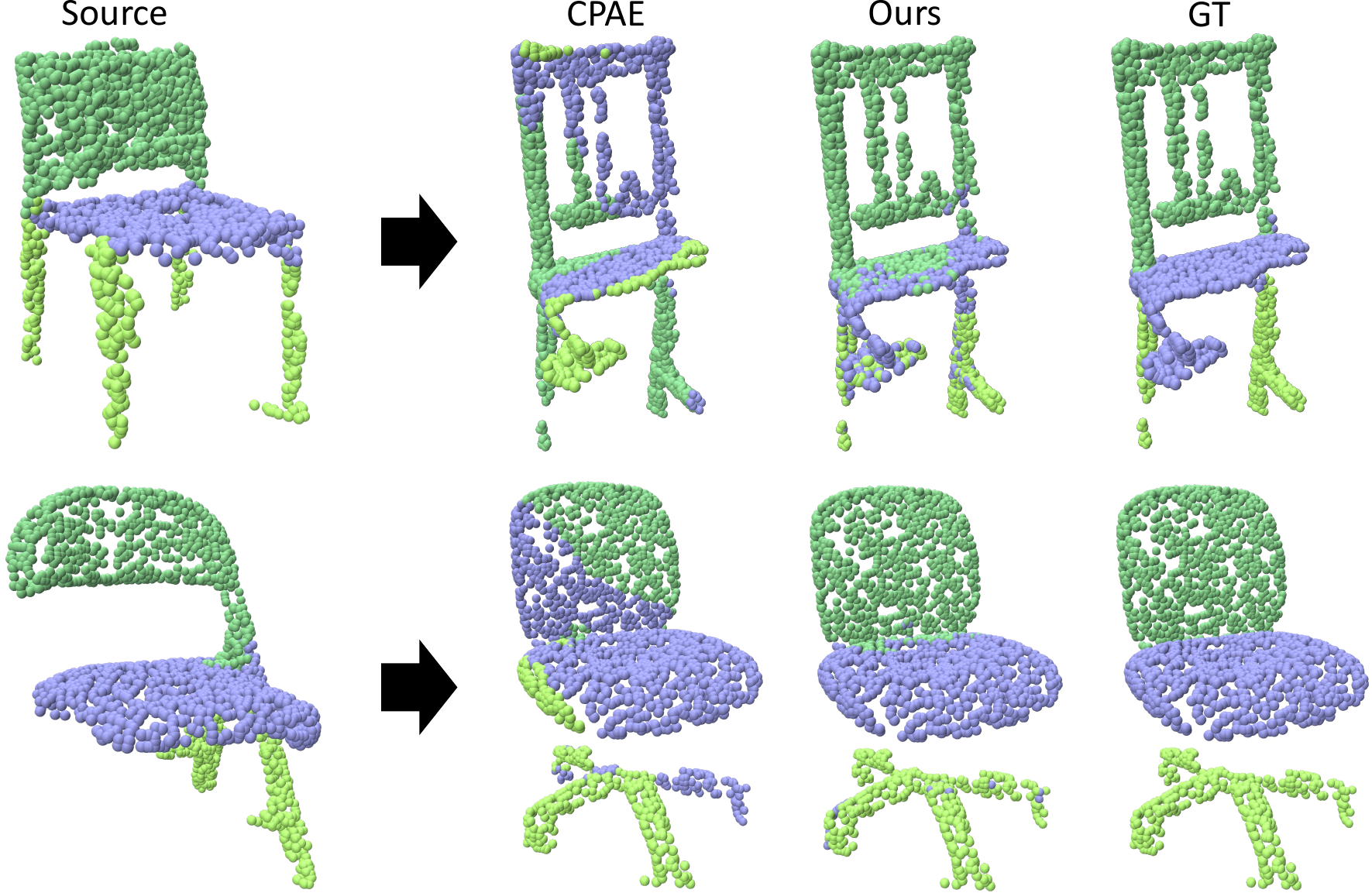}
    \caption{\textbf{Qualitative results of part label transfer on ScanObjectNN~\cite{scanobjectnn}.} Note that both source and target point clouds were arbitrarily rotated at evaluation, but have been aligned in the figure for better visibility of part label transfer results. The results show that \methodName reasonably predicts the semantic correspondences between arbitrarily rotated and partial real point clouds.}
    \vspace{-2mm}
    \label{fig:qual_scanobjectnn}
\end{figure}

\noindent \textbf{ScanObjectNN~\cite{scanobjectnn}.}
We evaluate each method trained on synthetic chair data of ShapeNet~\cite{chang2015shapenet} on the real chair data of ScanObjectNN~\cite{scanobjectnn}, which is partial and more challenging than ShapeNetPart~\cite{shapenetpart}, without fine-tuning.
As shown in Table~\ref{tbl:part_label_transfer_scanobjectnn}, \methodName consistently outperforms previous approaches~\cite{yang2018foldingnet, deprelle2019learning, lang2021dpc, cheng2021learning} both with and without rotation augmentations during training.
The qualitative results presented in Figure~\ref{fig:qual_scanobjectnn} show that \methodName can predict rotation-robust 3D semantic correspondence between real and partial chair shapes, while CPAE~\cite{cheng2021learning} fails.


\begin{figure*}[t]
    \begin{center}
        \includegraphics[width=1.0\linewidth]{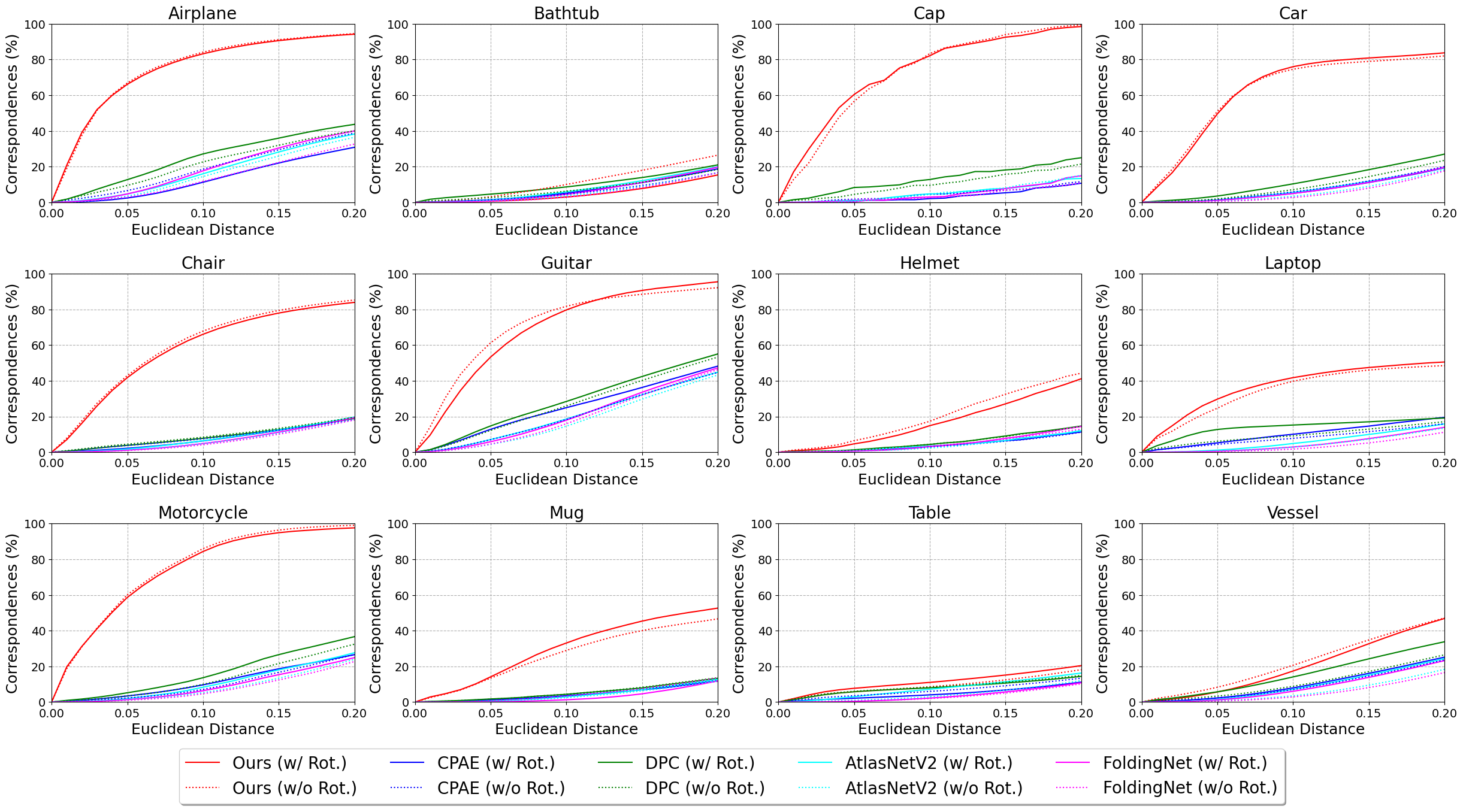}
    \end{center}
      \caption{\textbf{Percentage of Correct Keypoints (PCK) for the 12 categories of the KeypointNet dataset~\cite{you2020keypointnet} with and without rotation augmentations during training}.
      \methodName consistently outperforms previous approaches on all classes and thresholds in both settings.
}
\vspace{-2mm}
\label{fig:pck_keypointnet}
\end{figure*}

\subsection{3D Semantic Keypoint Transfer}
\begin{figure*}[t]
    \begin{center}
        \includegraphics[width=1.0\linewidth]{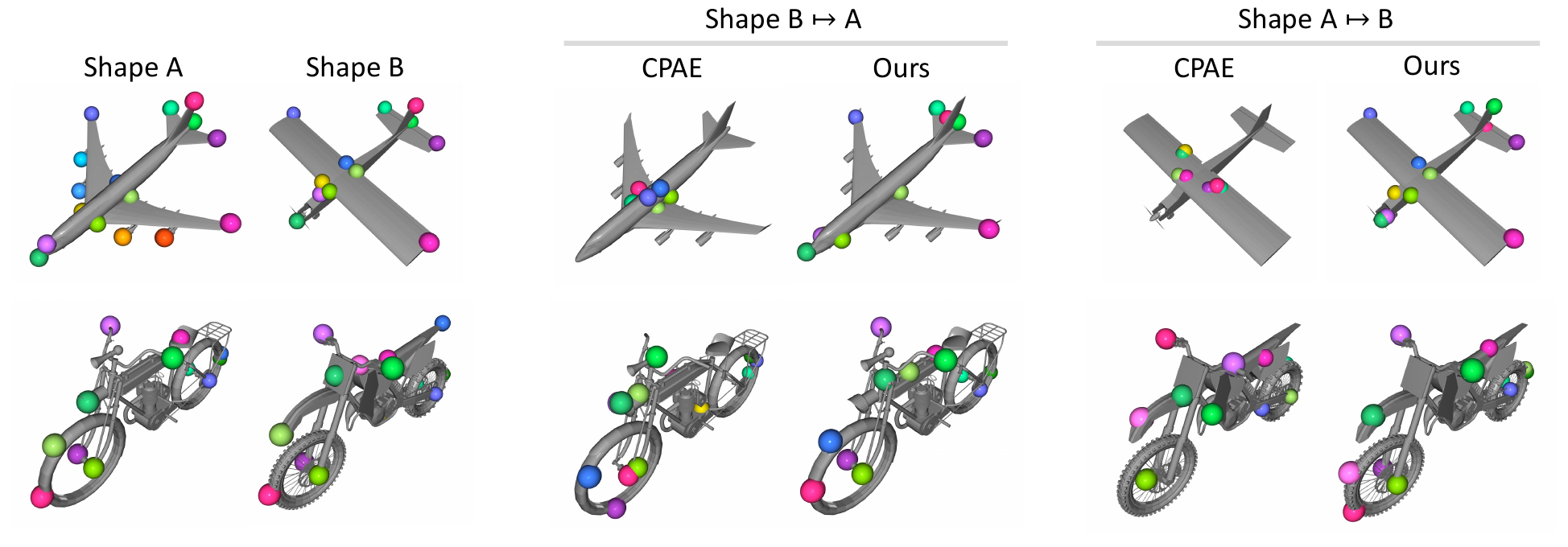}
    \end{center}
      \vspace{-2mm}
      \caption{\textbf{Keypoint transfer results for airplane and motorcycle categories of KeypointNet~\cite{you2020keypointnet}}. Each row contains a shape pair, each with ground-truth keypoints and the keypoint transfer results.
      Note that the input shapes were arbitrarily rotated at evaluation, but have been aligned in the above figure for better visibility of keypoint transfer results.
      \methodName shows to transfer the keypoints more accurately.
}
\vspace{-2mm}
\label{fig:qual_keypointnet}
\end{figure*}

Following the previous work~\citep{cheng2021learning}, we compute the distances from the transferred $M$ keypoints to the ground truth keypoints, and report PCK (Percentage of Correct Keypoints) of our transferred keypoints, which is computed by:
\begin{align}
\text{PCK} = \frac{1}{M}\sum_{m=1}^M\mathbbm{1}[\|\mathbf{k}_m - \hat{\mathbf{k}}_m\| \leq \tau],
\end{align}
where $\tau$, $\mathbf{k}_m$, and $\hat{\mathbf{k}}_m$ are a distance threshold, $m$-th ground truth keypoint, and $m$-th transferred keypoint, respectively.
The results on the KeypointNet dataset~\cite{you2020keypointnet} are illustrated in Figure~\ref{fig:pck_keypointnet} for varying distance thresholds $\tau$.
It can be seen that \methodName consistently outperforms baseline methods with and without rotation augmentations during training for all classes, by up to 10$\times$ on certain classes and thresholds.
This substantiates \methodName's superior efficacy at establishing dense 3D correspondences between varying shapes.
However, for certain classes such as Bathtub or Table, the performance is noticeably low, outperforming baseline methods only by a tight margin.
We speculate this to be due to the prevalent rotational symmetry of those classes, making it especially challenging to establish accurate 3D correspondences under arbitrary rotations.
The qualitative results of \methodName in comparison to baseline methods are presented in Figure~\ref{fig:qual_keypointnet}.
It can also be seen that \methodName can identify more accurate keypoint correspondences compared to CPAE~\cite{cheng2021learning} under arbitrary rotations, confirming the results of Figure~\ref{fig:pck_keypointnet}.


\begin{figure*}[t]
    \begin{center}
        \includegraphics[width=\linewidth]{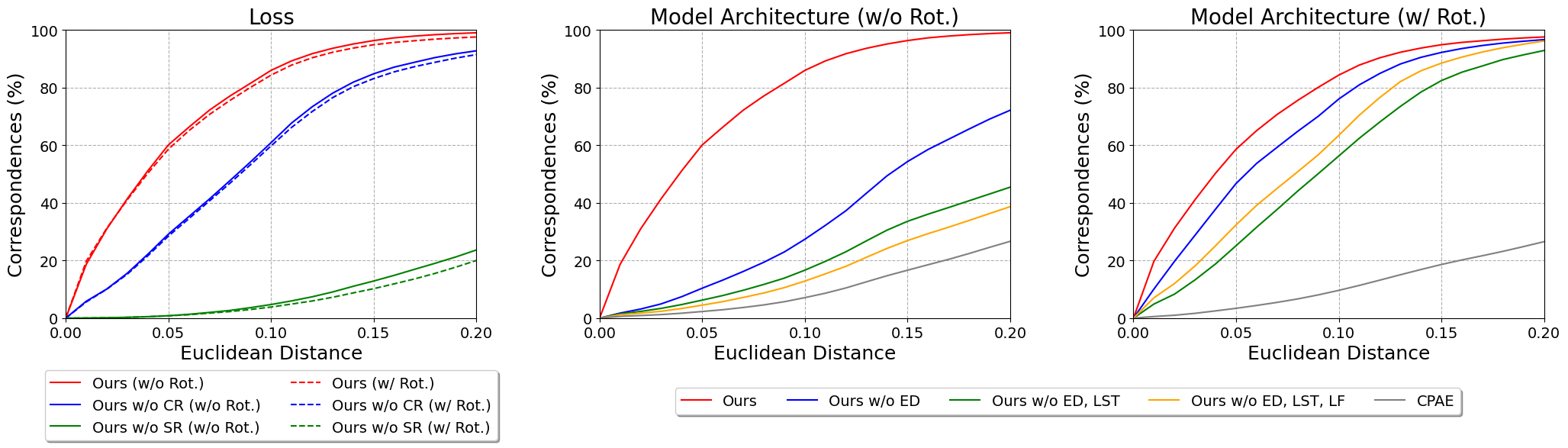}
    \end{center}
    \vspace{-2mm}
      \caption{\textbf{Ablation study on losses (the leftmost) and the components of model architecture (the others)}; self-reconstruction loss (SR), cross-reconstruction loss (CR), equivariant decoder (ED), local shape transform (LST), and local feature (LF).
      When excluding the equivariant decoder, local shape transforms, or local features, the default is to use an SO(3)-variant decoder, UV coordinates~\citep{cheng2021learning}, or global features for encoding, respectively.
}
\vspace{-2mm}
\label{fig:ablation}
\end{figure*}

\subsection{Ablation Study and Analyses}
We perform an ablation study to justify the design choice of \methodName, and evidence the efficacy of each component.

\smallbreak
\noindent
\textbf{Self- and cross-reconstruction.}
We train \methodName in a self-supervised manner via penalizing errors in self- and cross- reconstruction of input point clouds.
We conduct an ablation study on \methodName's reconstruction, providing comparative results for scenarios with and without its use.
The results are illustrated in the first graph of Figure~\ref{fig:ablation}.
It can be seen that with and without rotation augmentations during training, incorporating \textit{both} self- and cross-reconstruction yields the best results.
Removing self-reconstruction results in a much dramatic drop in performance; we conjecture this is because without self-reconstruction, the dynamic local shape transform (Sec.~\ref{subsubsec:encoder}) fails to capture the required locality of its own point cloud in the first place, being unsuitable to establish correspondences.

\smallbreak
\noindent
\textbf{Encoder outputs and SO(3)-equivariance.}
\methodName uses VNNs~\cite{deng2021vector} as the SO(3)-equivariant layers to facilitate 3D dense correspondence establishment between arbitrarily rotated point cloud pairs, leveraging local shape transform to map global shape descriptors to local shape descriptors which encode the pointwise semantics and local geometry.
We perform an ablation study to demonstrate the efficacy of local shape transforms and SO(3)-equivariant and -invariant representations in \methodName on the motorcycle class of the KeypointNet dataset~\cite{you2020keypointnet}.
We start our comparison from the architecture of CPAE~\citep{cheng2021learning}, given that they also employ an encoder-decoder architecture to self-supervise their network via shape reconstruction.
The results are presented in the two rightmost graphs of Figure~\ref{fig:ablation}, showing the evaluation results with and without rotation augmentations during training in order.
It can be seen that our design choice of using equivariant encoders and decoders shows consistent improvements over using an SO(3)-variant counterpart. 
Also, using UV coordinates as proposed in CPAE~\cite{cheng2021learning} performs worse compared to our dynamic local shape transform, evidencing the comparatively better efficacy of transforming each point to their local shape descriptors via our dynamic SO(3)-invariant shape transform.
While using local features as inputs to the encoder shows varied trends across with and without augmentation settings, using the point-wise local feature as input is a key component that facilitates the learning of \textit{point-wise} dynamic local shape transform that is essential in establishing the 3D correspondences in \methodName.

\section{Conclusion}
\label{sec:conclusion}


We've introduced \methodName, a novel self-supervised learner for dense 3D semantic matching across shapes of the same category, even with arbitrary rotations. Its robustness stems from our innovative use of SO(3)-equivariant and -invariant representations, enabling dynamic local shape transforms that preserve rotation equivariance. These transforms map global descriptors to local ones, facilitating the establishment of dense correspondences. Our method outperforms existing ones on tasks like part label and keypoint transfer, enhancing applicability in computer vision and robotics, e.g., AR/VR and texture mapping. Future research could focus on improving robustness under common corruption.

\vspace{2mm}
{
\small
\noindent\textbf{Acknowledgement.}
This work was supported by IITP grants (No.2021-0-02068: AI Innovation Hub (50\%), No. 2022-0-00290: Visual Intelligence for Space-Time Understanding and Generation (40\%), No.2019-0-01906: Artificial Intelligence Graduate School Program at POSTECH (5\%), NO.2021-0-01343, Artificial Intelligence Graduate School Program at Seoul National University (5\%)) funded by the Korea government (MSIT).
Seungwook was supported by the Hyundai-Motor Chung Mong-koo Foundation.
}


\clearpage
\appendix
\setcounter{table}{0}
\renewcommand{\thetable}{A\arabic{table}}%
\setcounter{figure}{0}
\renewcommand{\thefigure}{A\arabic{figure}}%
\maketitlesupplementary

In this supplementary material, we provide a detailed explanation of \methodName and additional experiment results.

\section{Inference Algorithm of \methodName}
\label{sec:supp_algo}
In this section, we provide a detailed algorithm for the inference process of \methodName for a better understanding of \methodName.
As shown in Algorithm~\ref{alg:inference}, given a pair of shapes, we cross-reconstruct one shape and find the nearest neighbor for each point of the cross-reconstructed shape from the other shape, following the inference algorithm of the previous work~\cite{cheng2021learning}.

\begin{algorithm}[h]
    \small
    \caption{: Inference}
    \label{alg:inference}
    \textbf{Input:} A pair of shapes $(\mathbf{P}, \mathbf{Q})$, $\operatorname{Encoder}(\cdot)$, $\operatorname{Decoder}(\cdot)$
    
    \textbf{Output:} Correspondence $\mathcal{C}$ for all $\mathbf{p}\in\mathbf{P}$
    
    \begin{algorithmic}[1]
    
    \State $\mathbf{Z}_p$, $\{f^p_{\theta_i}\}$ $\leftarrow$ $\operatorname{Encoder}(\mathbf{P})$
    \State $\mathbf{Z}_q$, $\{f^q_{\theta_i}\}$ $\leftarrow$ $\operatorname{Encoder}(\mathbf{Q})$
    \State $\mathbf{Q}'$ $\leftarrow$ $\operatorname{Decoder}(\{f^p_{\theta_i}(\mathbf{Z}_q)\})$ \Comment{Cross-recon. from $\mathbf{P}$ to $\mathbf{Q}$}
    \State $\mathcal{C} \leftarrow \{\}$ \Comment{Initialization}
    \For{\{{$i \leftarrow 1$ to $|\mathbf{P}|$}\}}
        \State $\mathbf{p} \leftarrow \mathbf{P}_i$
        \State $\mathbf{q} \leftarrow \operatorname{NearestNeighborSearch}(\mathbf{Q}'_i, \mathbf{Q})$
        \State $\mathcal{C} \leftarrow \mathcal{C}~\cup~(\mathbf{p}, \mathbf{q})$
    \EndFor
    \end{algorithmic}
\end{algorithm}
\section{Ablation Study on Losses}
In this section, we conduct an ablation study of the components of our self-reconstruction loss (Eq.~\ref{eq:sr}) on the motorcycle category of the ShapeNetPart dataset~\cite{shapenetpart}.
As shown in Table~\ref{tbl:supp_loss}, our choice (b) shows the best performance among models trained with the seven loss variants from (a) to (g).

\begin{table}[h]
    \centering
    \begin{adjustbox}{width=0.65\linewidth,center}
        \begin{tabular}{ccccc}
            \toprule
            & $\operatorname{MSE}$ & $\operatorname{EMD}$ & $\operatorname{CD}$ & IoU ($\%$) \\
            \midrule
            (a) & \cmark & \cmark & \cmark & 46.0 \\
            (b) & \cmark & \cmark & & \textbf{48.5} \\
            (c) & \cmark & & \cmark & \underline{47.0} \\
            (d) & & \cmark & \cmark & 46.4 \\
            (e) & & & \cmark & 45.8 \\
            (f) & & \cmark & & 46.4 \\
            (g) & \cmark & & & 44.9 \\
            \bottomrule
        \end{tabular}
    \end{adjustbox}
    \caption{
\textbf{Ablation study on the self-reconstruction loss.} The model trained with ((b): $\operatorname{MSE}$ and $\operatorname{EMD}$) shows the best performance, justifying our choice for the self-reconstruction loss.
}
    \label{tbl:supp_loss}
\end{table}




\section{Multi-class Training}
In this section, we provide the experiment results of previous approaches~\cite{yang2018foldingnet, deprelle2019learning, lang2021dpc, cheng2021learning} and ours trained with multiple classes (airplane and chair) in the ShapeNetPart dataset~\cite{shapenetpart}. As shown in Table~\ref{tbl:supp_multi}, \methodName outperforms previous approaches~\cite{yang2018foldingnet, deprelle2019learning, lang2021dpc, cheng2021learning} by a large margin.
\begin{table}[h]
    \centering
    \begin{adjustbox}{width=0.75\linewidth,center}
        \begin{tabular}{lccc}
            \toprule
            Method & Airplane & Chair & Average \\
            \midrule
            FoldingNet~\cite{yang2018foldingnet} & 20.9 & 23.9 & 22.4 \\
            AtlasNetV2~\cite{deprelle2019learning} & 21.1 & 24.6 & 22.9 \\
            DPC~\cite{lang2021dpc} & 22.7 & 25.6 & \underline{24.2} \\
            CPAE~\cite{cheng2021learning} & 16.6 & 14.8 & 15.7 \\
            \methodName (ours) & 34.4 & 34.7 & \textbf{34.6} \\
            \bottomrule
        \end{tabular}
    \end{adjustbox}
    \caption{
\textbf{Part label transfer with multi-classes training.} 
}
    \label{tbl:supp_multi}
\end{table}
\section{Generalization to Unseen Classes}
In this section, we evaluate the generalization ability of previous approaches~\cite{yang2018foldingnet, deprelle2019learning, lang2021dpc, cheng2021learning} and ours to unseen classes. Specifically, we train each method on the airplane category in the ShapeNetPart dataset~\cite{shapenetpart} and test it on the chair category. As shown in Table~\ref{tbl:supp_unseen}, \methodName shows a competitive result with an unseen category, outperforming previous approaches~\cite{yang2018foldingnet, deprelle2019learning, cheng2021learning} except DPC~\cite{lang2021dpc}.
\begin{table}[h]
    \centering
    \begin{adjustbox}{width=\linewidth,center}
       \begin{tabular}{ccccc}
                \toprule
                FoldingNet~\cite{yang2018foldingnet} & AtlasNetV2~\cite{deprelle2019learning} & DPC~\cite{lang2021dpc} & CPAE~\cite{cheng2021learning} & \methodName \\                
                \midrule
                24.8 & 23.0 & \textbf{28.2} & 15.6 & \underline{27.3} \\
                \bottomrule
        \end{tabular}
        \end{adjustbox}
        \caption{\textbf{Generalization results for the part label transfer.}
        }
        \label{tbl:supp_unseen}
\end{table}
\section{Inference on Aligned Shapes}
\begin{figure*}[h]
    \begin{center}
        \includegraphics[width=1.0\linewidth]{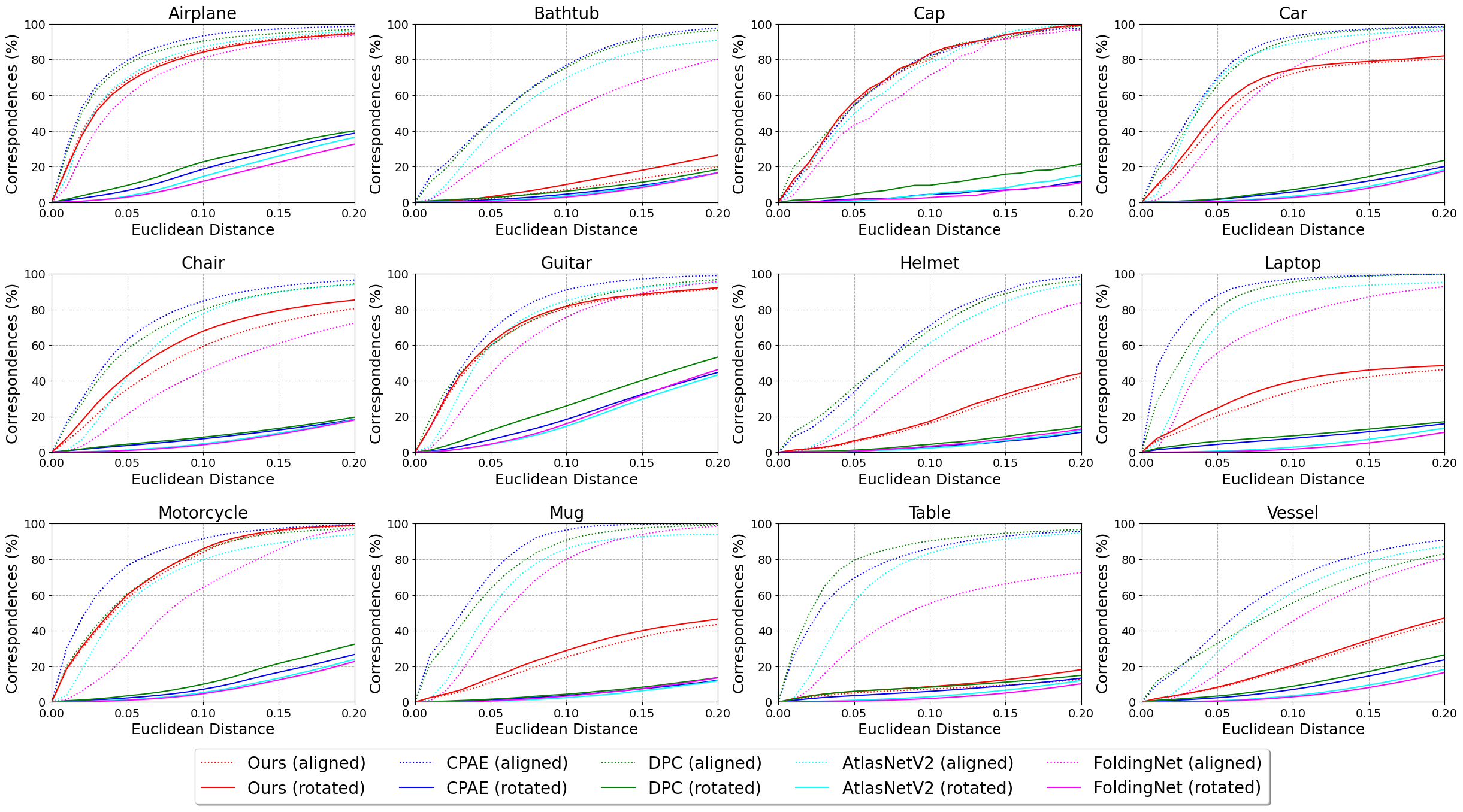}
    \end{center}
      \caption{\textbf{Percentage of Correct Keypoints (PCK) for the 12 categories of the KeypointNet dataset~\cite{you2020keypointnet} on aligned and rotated shapes.} Note that each method is trained without rotation augmentation. While previous approaches~\cite{yang2018foldingnet, deprelle2019learning, lang2021dpc, cheng2021learning} are vulnerable to rotations, \methodName shows a negligible performance drop with rotated shapes.
}
\label{fig:supp_kpn_aligned}
\end{figure*}

In this section, we provide the results of previous approaches~\cite{yang2018foldingnet, deprelle2019learning, lang2021dpc, cheng2021learning} and ours evaluated on the ShapeNetPart~\cite{shapenetpart}, ScanObjectNN~\cite{scanobjectnn}, and KeypointNet~\cite{you2020keypointnet} datasets, but with aligned test shapes, as shown in Table~\ref{tbl:supp_part_label_transfer}, Table~\ref{tbl:supp_part_label_transfer_scanobjectnn}, and Figure~\ref{fig:supp_kpn_aligned}, respectively.
Note that under the aligned setting, the input shape pairs are perfectly aligned both at train and test time - which is an unrealistic setting in practice.
For each method, we also include the results with rotated shapes to show the performance difference between aligned and rotated settings.
It can be seen that while the drop in performance for previous approaches~\cite{yang2018foldingnet, deprelle2019learning, lang2021dpc, cheng2021learning} from the aligned to the rotated setting is drastic, the difference is negligible in \methodName, demonstrating the robustness of our SO(3) correspondence establishment scheme against arbitrary rotations.
While \methodName is not always competitive on all settings, it is impractical to expect perfectly aligned shapes in real-world situations; on the realistic setting of SO(3) evaluation, \methodName consistently shows the best results.

\clearpage
\begin{table*}[h]
    \centering
    \begin{adjustbox}{width=0.95\textwidth,center}
       \begin{tabular}{clccccccccc}
                \toprule
                Inference & Method & Airplane & Cap & Chair & Guitar & Laptop & Motorcycle & Mug & Table & Average \\                
                \midrule
                \multirow{5}{*}[-0.5ex]{\shortstack{Aligned}} & FoldingNet~\cite{yang2018foldingnet}
                                   & 56.5 & 54.9 & 63.1 & 73.1 & 81.9 & 21.5 & 75.5 & 54.0 & 60.1 \\
                & AtlasNetV2~\cite{deprelle2019learning} 
                                   & 51.7 & 44.7 & 63.3 & 65.0 & 84.0 & 41.5 & 84.2 & 59.3 & 61.7 \\
                & DPC~\cite{lang2021dpc} 
                                   & 60.5 & 65.8 & 65.3 & 74.4 & 88.0 & 53.3 & 85.4 & 66.4 & \underline{69.9} \\
                & CPAE~\cite{cheng2021learning} 
                                   & 61.3 & 61.6 & 72.6 & 78.9 & 89.9 & 55.4 & 86.5 & 72.5 & \textbf{72.3} \\
                & \methodName (ours) & 52.1 & 54.4 & 58.3 & 74.1 & 56.7 & 48.7 & 75.6 & 41.3 & 57.7 \\
                \midrule
                \multirow{5}{*}[-0.5ex]{\shortstack{Rotated}} & FoldingNet~\cite{yang2018foldingnet}
                                   & 17.8 & 34.7 & 22.5 & 22.1 & 36.2 & 12.6 & 50.0 & 34.6 & 28.8 (\textcolor{purple}{$\downarrow$ 31.3}) \\
                & AtlasNetV2~\cite{deprelle2019learning} 
                                   & 19.7 & 31.4 & 23.6 & 22.7 & 36.0 & 13.1 & 49.7 & 35.2 & 28.9 (\textcolor{purple}{$\downarrow$ 32.8}) \\
                & DPC~\cite{lang2021dpc} 
                                   & 22.7 & 37.1 & 25.6 & 31.9 & 35.0 & 17.5 & 51.3 & 36.8 & \underline{32.2} (\textcolor{purple}{$\downarrow$ 37.7}) \\
                & CPAE~\cite{cheng2021learning} 
                                   & 21.0 & 38.0 & 26.0 & 22.7 & 34.9 & 14.7 & 51.4 & 35.5 & 30.5 (\textcolor{purple}{$\downarrow$ 41.8}) \\
                & \methodName (ours) & 52.1 & 54.5 & 58.3 & 74.1 & 56.5 & 48.6 & 75.0 & 41.3 & \textbf{57.6} (\textcolor{purple}{$\downarrow$ ~~0.1}) \\
               
                \bottomrule
        \end{tabular}
        \end{adjustbox}
        \caption{\textbf{Average IoU (\%) of part label transfer for eight categories in the ShapeNetPart dataset~\cite{shapenetpart} on aligned and rotated shapes.} 
        Note that each method is trained without rotation augmentation. \methodName shows the most negligible performance drop ($0.1\%$ in IoU) with rotated shapes, while previous approaches~\cite{yang2018foldingnet, deprelle2019learning, lang2021dpc, cheng2021learning} show large performance drops (at least $30\%$ in IoU).
        }
        \label{tbl:supp_part_label_transfer}
\end{table*}

\begin{table*}[h]
    \centering
    \begin{adjustbox}{width=0.72\textwidth,center}
       \begin{tabular}{cccccc}
                \toprule
                Inference & FoldingNet~\cite{yang2018foldingnet} & AtlasNetV2~\cite{deprelle2019learning} & DPC~\cite{lang2021dpc} & CPAE~\cite{cheng2021learning} & \methodName (ours) \\                
                \midrule
                Aligned & 33.6 & 34.8 & \underline{36.3} & 33.8 & \textbf{39.6} \\
                \midrule
                Rotated & 23.2 (\textcolor{purple}{$\downarrow$ 10.4}) & 23.6 (\textcolor{purple}{$\downarrow$ 11.2}) & 23.9 (\textcolor{purple}{$\downarrow$ 12.4}) & \underline{24.4} (\textcolor{purple}{$\downarrow$ 9.4}) & \textbf{39.6} (\textcolor{teal}{$-$}) \\
                \bottomrule
        \end{tabular}
        \end{adjustbox}
        \caption{\textbf{Average IoU (\%) of part label transfer for the chair category in the ScanObjectNN dataset~\cite{scanobjectnn} on aligned and rotated shapes.} 
        Note that each method is trained without rotation augmentation. \methodName does not show any performance drop with rotated shapes, while previous approaches~\cite{yang2018foldingnet, deprelle2019learning, lang2021dpc, cheng2021learning} show large performance drops (at least $9\%$ in IoU).
        }
        \label{tbl:supp_part_label_transfer_scanobjectnn}
\end{table*}
\clearpage

\begin{figure}[t]
    \begin{center}
        \includegraphics[width=\linewidth]{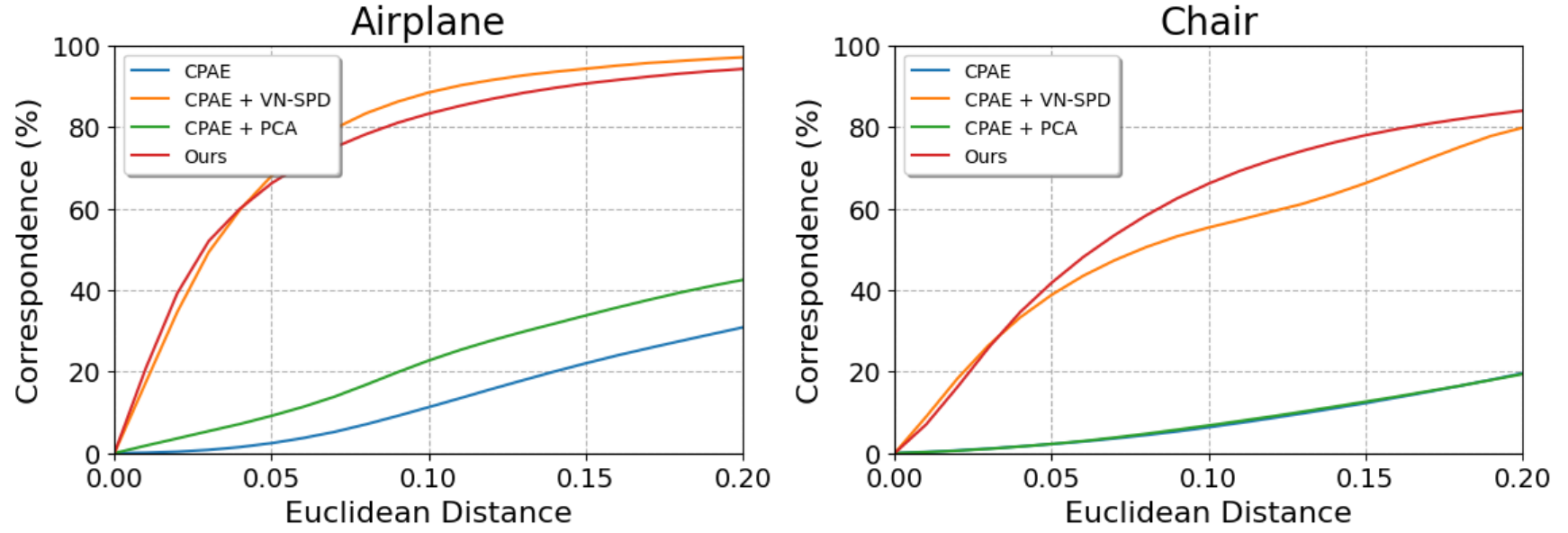}
    \end{center}
    \caption{\textbf{Comparision of RIST with the combination of CPAE~\cite{cheng2021learning} and aligning methods; PCA and VN-SPD~\cite{katzir2022shape}.} Note that RIST shows competitive results to the combined method of CPAE~\cite{cheng2021learning} and VN-SPD~\cite{katzir2022shape}, which is a SE(3)-equivariant orientation predictor and requires additional parameters (17.4M).}
\label{fig:supp_aligner}
\end{figure}

We further evaluate CPAE~\cite{cheng2021learning} when integrated together with two 3D shape alignment methods.
We present results when using PCA or VN-SPD~\cite{katzir2022shape} as the alignment method in Figure~\ref{fig:supp_aligner}.
PCA yields only marginal performance improvements, likely due to sign and order ambiguities. In contrast, using the SOTA learning-based alignment method, VN-SPD, produces competitive results with \methodName, but its inconsistency in aligning shapes to a canonical orientation limits performance as also reported in Katzir \etal~\cite{katzir2022shape}, sacrificing the efficiency.

\section{Comparision with 3D Keypoint Estimators}
We compare \methodName with SC3K~\cite{zohaib2023sc3k}, a recent self-supervised method for coherent 3D keypoint estimation. However, since this 3D keypoint \textit{estimation} method cannot be evaluated on 3D keypoint \textit{matching}, we instead used the Dual Alignment Score (DAS)\footnote{A metric for 3D keypoint estimation task SC3K~\cite{zohaib2023sc3k} used.} of RIST for an empirical comparison with SC3K.
Additionally, to facilitate comparison on the part label transfer task, we extended the number of keypoints estimated by SC3K to match the total number of points in a point cloud \eg, 2048.

\begin{table}[h]
    \centering
    \begin{adjustbox}{width=0.65\linewidth,center}
       \begin{tabular}{lccc}
                \toprule
                Method & Airplane & Car & Chair \\
                \midrule
                SC3K~\cite{zohaib2023sc3k} & 81.3 & 73.8 & \textbf{86.2} \\
                RIST (ours) & \textbf{82.4} & \textbf{76.9} & 81.8 \\
                \bottomrule
        \end{tabular}
        \end{adjustbox}
        \caption{\textbf{Dual Alignment Score of SC3K~\cite{zohaib2023sc3k} and RIST.} During the evaluation, we use 10 keypoints for both SC3K and RIST.
        }
        \label{tbl:supp_sc3k}
\end{table}
\begin{table}[h]
    \centering
    \begin{adjustbox}{width=0.65\linewidth,center}
       \begin{tabular}{lccc}
                \toprule
                Method & Airplane & Car & Chair \\
                \midrule
                SC3K~\cite{zohaib2023sc3k} & 22.3 & 23.0 & 24.7 \\
                RIST (ours) & \textbf{51.2} & \textbf{48.0} & \textbf{55.0} \\
                \bottomrule
        \end{tabular}
        \end{adjustbox}
        \caption{\textbf{Part label transfer results of SC3K~\cite{zohaib2023sc3k} and RIST.} Note that we train SC3K~\cite{zohaib2023sc3k} with 2048 keypoints.
        }
        \label{tbl:supp_sc3k_2048}
\end{table}

As shown in Tables~\ref{tbl:supp_sc3k} and~\ref{tbl:supp_sc3k_2048}, RIST exhibits competitive DAS results compared to SC3K, although it is not trained for 3D keypoint estimation, and significantly outperforms SC3K on the part label transfer task.
\section{Evaluation with Pseudo-Ground Truth}
\begin{figure*}[t]
    \centering
    \includegraphics[width=\linewidth]{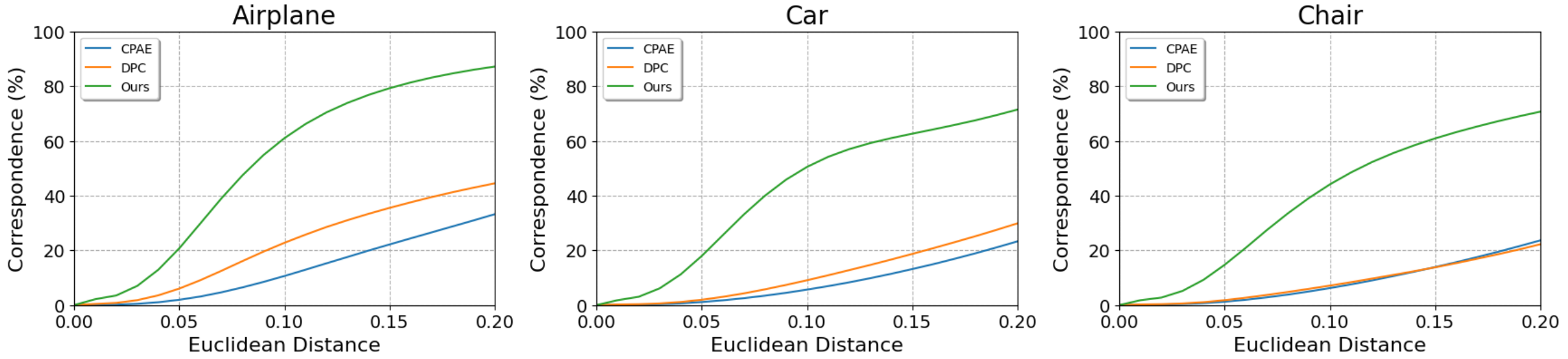}
    \caption{\textbf{3D semantic correspondence results of RIST using DIT~\cite{zheng2021deep} as pseudo-ground truth on ShapeNet~\cite{chang2015shapenet}}. We use the official checkpoints of DIT to generate pseudo-ground truth of 3D semantic correspondence for rotated 3D shapes.}
    \label{fig:rebuttal_dit}
\end{figure*}
We utilize DIT~\cite{zheng2021deep} to establish pseudo-ground truth on ShapeNet~\cite{chang2015shapenet} for a direct evaluation of \methodName's dense semantic correspondence capabilities for airplane, car, and chair classes, using official checkpoints.
As shown in Figure~\ref{fig:rebuttal_dit}, the results show a similar trend of part label transfer results, showing that RIST outperforms previous approaches.
\section{Implementation Details of SO(3)}
\label{sec:supp_so3}
In this section, we explain the implementation details of uniformly sampling random rotations and highlight the differences from the previous approach~\cite{cheng2021learning}. Cheng~\etal~\cite{cheng2021learning} samples rotation angles from $\mathcal{N}(0, 0.2^2)$ and then clamps them to $[-\frac{1}{2}\pi,\frac{1}{2}\pi]$, which limits the range of rotation. In our work, we follow Shoemake \etal~\cite{shoemake1992uniform} to \textit{uniformly} sample to cover full SO(3), which is more challenging.
\section{Part Label Transfer Results on More Classes}
\label{sec:supp_partlabeltransfer}
We initially presented evaluations only for the classes of ShapeNetPart~\cite{shapenetpart} that are shared with those of KeypointNet~\cite{you2020keypointnet}. In Table~\ref{tbl:rebuttal_shapenetpart}, we further present part label transfer results on the remaining classes of ShapeNetPart~\cite{shapenetpart}.

\begin{table}[h]
    \centering
    \begin{adjustbox}{width=\linewidth,center}
       \begin{tabular}{lcccccccc}
                \toprule
                Method & Bag & Car & Ear. & Knife & Lamp & Pistol & Rocket & Skate. \\
                \midrule
                CPAE~\cite{liu2020learning} & 43.2 & 20.3 & 33.4 & 36.3 & 31.1 & 26.8 & 27.7 & 52.0 \\
                RIST (ours) & \textbf{50.8} & \textbf{48.0} & \textbf{36.3} & \textbf{57.9} & \textbf{35.9} & \textbf{54.7} & \textbf{34.4} & \textbf{54.4} \\
                \bottomrule
        \end{tabular}
        \end{adjustbox}
        \caption{\textbf{Part label transfer results on ShapeNetPart~\cite{shapenetpart}.} RIST consistently outperforms the previous state-of-the-art method on the remaining classes of ShapeNetPart~\cite{shapenetpart}.}
        \label{tbl:rebuttal_shapenetpart}
\end{table}
\section{Matching with Local Shape Transform}
In this section, We experiment with a variant of \methodName (RIST$_\text{LST}$), which matches 3D shapes using similarity between SO(3)-invariant Local Shape Transform (LST).
As shown in Table~\ref{tbl:rebuttal_matching_scheme}, our current scheme of comparing point positions yields noticeably better results on ShapeNetPart~\cite{shapenetpart}, meaning that our trained decoder is better adept at handling topology-varying structures

\begin{table}[t]
    \centering
    \begin{adjustbox}{width=0.45\linewidth,center}
       \begin{tabular}{lcc}
            \toprule
            Method & Airplane & Chair \\
            \midrule
            CPAE & 17.0 & 24.5 \\
            RIST$_\text{LST}$ & 48.6 & 50.3 \\
            RIST & \textbf{51.2} & \textbf{55.0} \\
            \bottomrule
        \end{tabular}
        \end{adjustbox}
        \caption{\textbf{Part label transfer results of RIST$_\text{LST}$.} Note that we use randomly rotated 3D shapes for the evaluation.}
        \label{tbl:rebuttal_matching_scheme}
\end{table}
\section{Alignment of Qualitative Results}
\begin{figure}[t]
    \begin{center}
        \includegraphics[width=\linewidth]{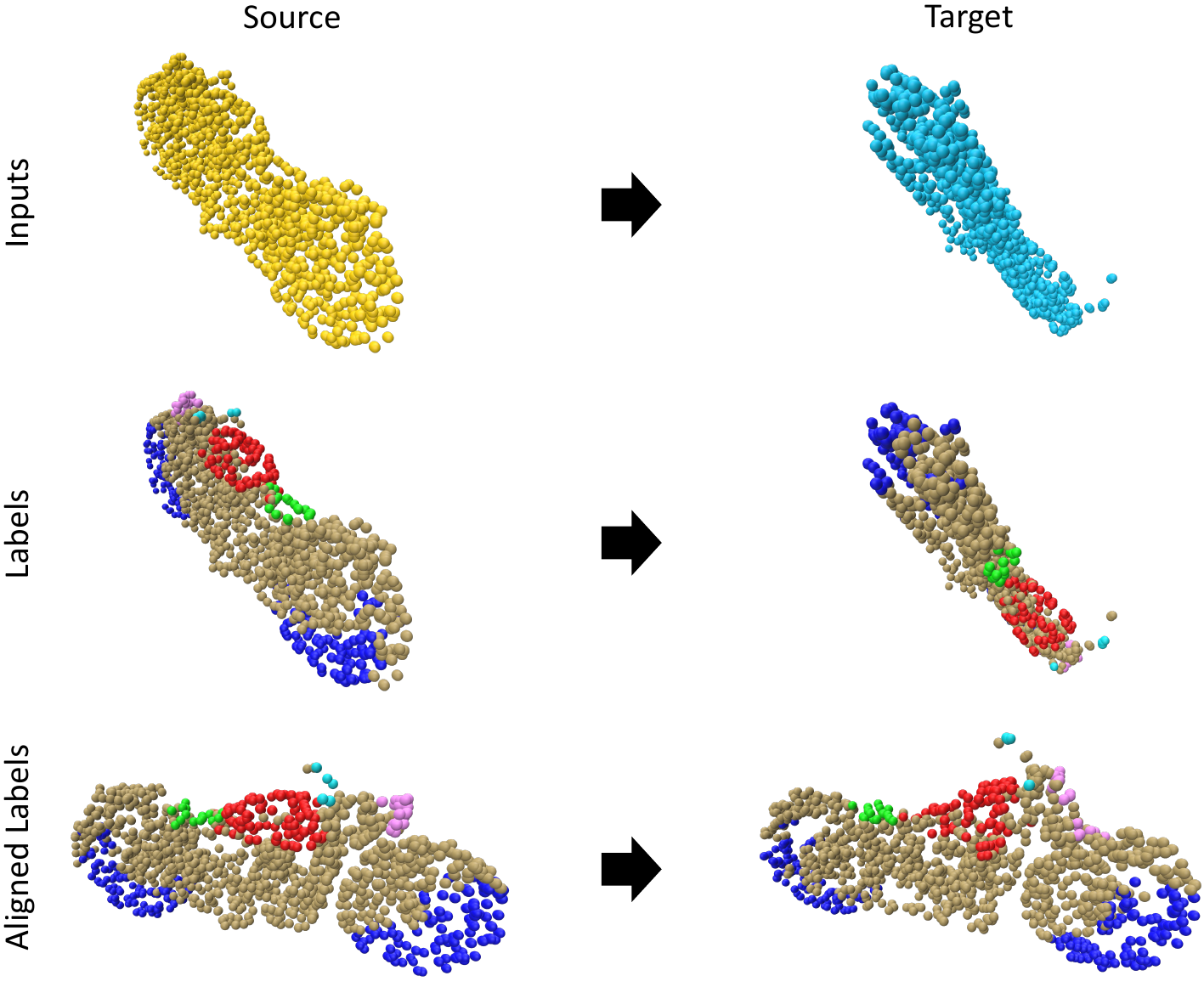}
    \end{center}
      \caption{\textbf{Visualization for aligning qualitative results of part label transfer on the ShapeNetPart dataset~\cite{shapenetpart}.}
}
\label{fig:supp_qual_alignment}
\end{figure}

In this section, we provide both unaligned and aligned qualitative results for a better understanding of how our qualitative results were drawn.
As shown in Figure~\ref{fig:supp_qual_alignment}, both source and target shapes are randomly rotated at the inference time.
Note that we use the part segmentation labels transferred by \methodName for the target shape in Figure~\ref{fig:supp_qual_alignment}.

\section{Qualitative Results of Part Label Transfer}
\begin{figure*}[h]
    \begin{center}
        \includegraphics[width=\linewidth]{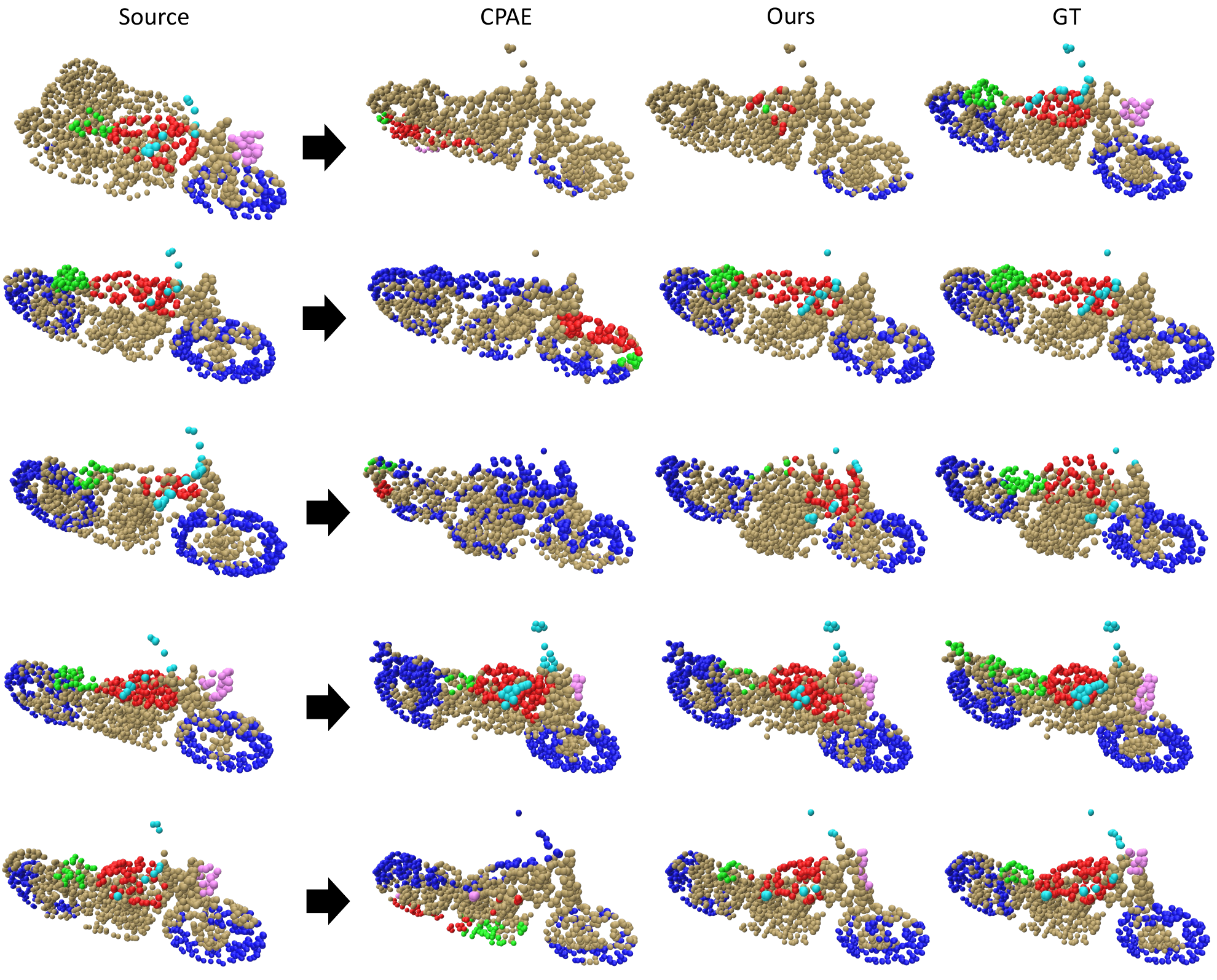}
    \end{center}
      \caption{\textbf{Qualitative results of part label transfer on the motorcycle class in the ShapeNet part dataset~\cite{shapenetpart}.} Note that the input shapes were arbitrarily rotated, differently for each target column, but have been aligned for better visibility of part label transfer results.
      \methodName shows to outperform CPAE~\cite{cheng2021learning} consistently, showing a high resemblance to ground truth results.
}
\label{fig:supp_part_motor}
\end{figure*}

\begin{figure*}[h]
    \begin{center}
        \includegraphics[width=\linewidth]{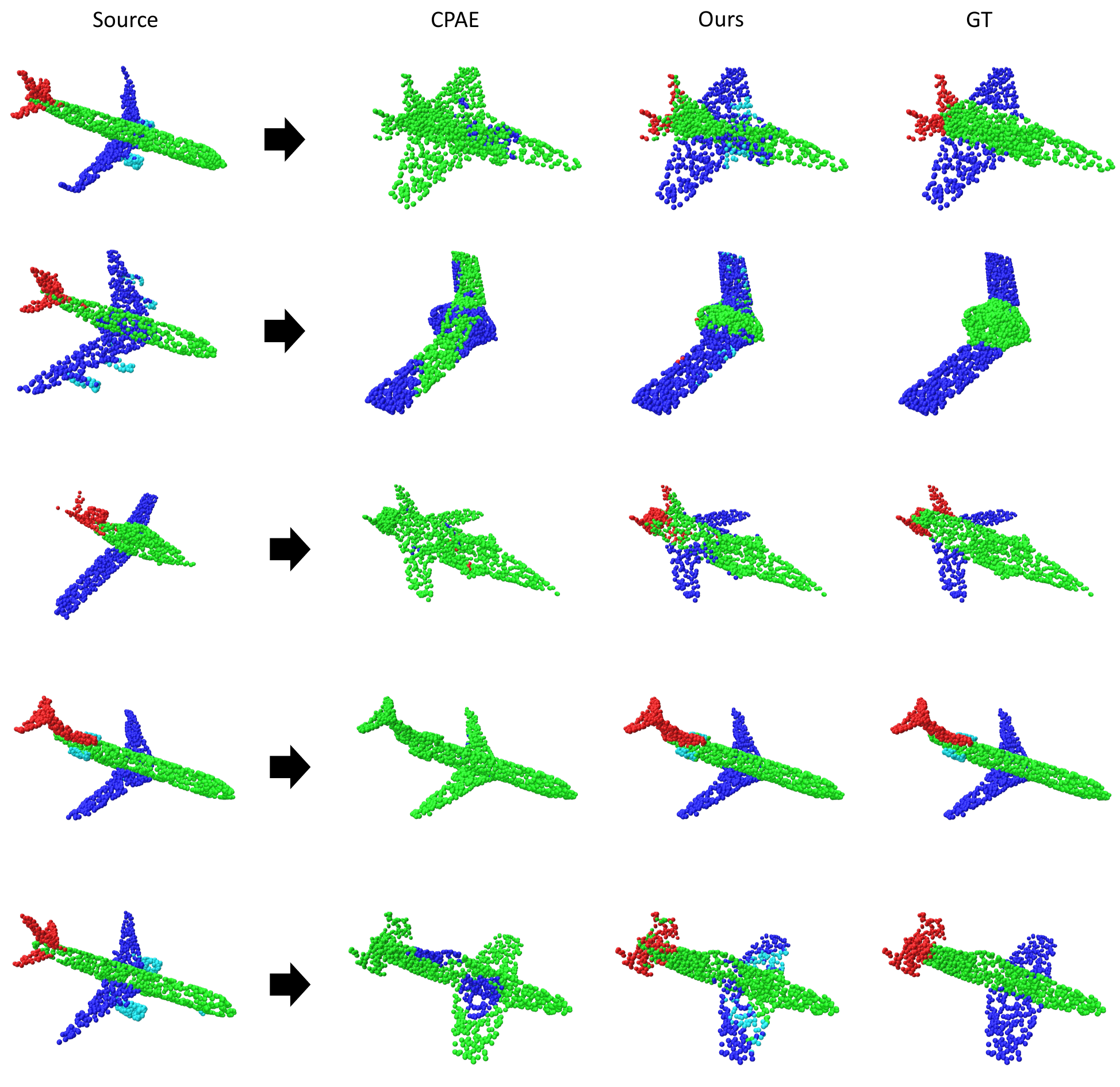}
    \end{center}
      \caption{\textbf{Qualitative results of part label transfer on the airplane class in the ShapeNet part dataset~\cite{shapenetpart}.} Note that the input shapes were arbitrarily rotated, differently for each target column, but have been aligned for better visibility of part label transfer results.
      \methodName shows to outperform CPAE~\cite{cheng2021learning} consistently, showing a high resemblance to ground truth results.
}
\label{fig:supp_part_airplane}
\end{figure*}

We provide additional qualitative results on the ShapeNetPart dataset~\cite{shapenetpart} that were not included in our manuscript due to space constraints, as shown in Figures~\ref{fig:supp_part_motor}~and~\ref{fig:supp_part_airplane}. 

\clearpage
\clearpage





%


\clearpage
{
    \small
    \bibliographystyle{iclr2024_conference}
    \bibliography{iclr2024_conference}
}


\end{document}